\documentclass[10pt,twocolumn,letterpaper]{article}

\usepackage{cvpr}              % To produce the CAMERA-READY version
\usepackage{graphicx}
\usepackage{amsmath}
\usepackage{amssymb}
\usepackage{booktabs}

\usepackage{algorithm}%算法
\usepackage{algorithmic}%算法

\usepackage{multirow, booktabs}
\usepackage{amssymb}%不规则表格
\usepackage{diagbox}%不规则表格

\usepackage[pagebackref,breaklinks,colorlinks]{hyperref}

\usepackage[capitalize]{cleveref}
\crefname{section}{Sec.}{Secs.}
\Crefname{section}{Section}{Sections}
\Crefname{table}{Table}{Tables}
\crefname{table}{Tab.}{Tabs.}

\def\cvprPaperID{7685} % *** Enter the CVPR Paper ID here
\def\confName{CVPR}
\def\confYear{2023}

\begin{document}

\title{Adversarial Color Film: Effective Physical-World Attack to DNNs}

\author{Chengyin Hu\\
University of Electronic Science \\and Technology of China\\
Chengdu, China\\
{\tt\small cyhuuestc@gmail.com}
\and
Weiwen Shi\\
University of Electronic Science \\and Technology of China\\
Chengdu, China\\
{\tt\small Weiwen\_shi@foxmail.com}
% \and
% Ling Tian \thanks{Corresponding author}\\
% University of \\Electronic Science and\\ Technology of China\\
% Chengdu, China\\
% {\tt\small lingtian@uestc.edu.cn}
}
\maketitle

%%%%%%%%% ABSTRACT
\begin{abstract}
It is well known that the performance of deep neural networks (DNNs) is susceptible to subtle interference. So far, camera-based physical adversarial attacks haven't gotten much attention, but it is the vacancy of physical attack. In this paper, we propose a simple and efficient camera-based physical attack called Adversarial Color Film (\textbf{AdvCF}), which manipulates the physical parameters of color film to perform attacks. Carefully designed experiments show the effectiveness of the proposed method in both digital and physical environments. In addition, experimental results show that the adversarial samples generated by AdvCF have excellent performance in attack transferability, which enables AdvCF effective black-box attacks. At the same time, we give the guidance of defense against AdvCF by means of adversarial training. Finally, we look into AdvCF's threat to future vision-based systems and propose some promising mentality for camera-based physical attacks.
\end{abstract}

%%%%%%%%% BODY TEXT
\section{Introduction}
\label{sec1}

Nowadays, vision-based systems and applications are gradually popularized in people's daily life, such as autonomous driving systems, unmanned aerial vehicles and so on. At the same time, the security and reliability of these systems are also the focus of many scholars. Most scholars are keen to study adversarial attacks in the digital environment \cite{ref18,ref19,ref20,ref21}, which fool advanced DNNs by adding carefully designed pixel-level adversarial perturbations to the input image, generating perturbations that are imperceptible to the human observers. In addition, some scholars are gradually working on the study of adversarial attacks in the physical environment \cite{ref22,ref23,ref24}, which uses stickers and graffiti as perturbations to fool advanced DNNs, generating perturbations visible to human observers. However, in the physical world, images are captured by the camera and then transmitted to the advanced DNNs, where an attacker cannot directly modify the input image.

\begin{figure}
\centering
\setlength{\belowcaptionskip}{-0.3cm}
\includegraphics[width=1\columnwidth]{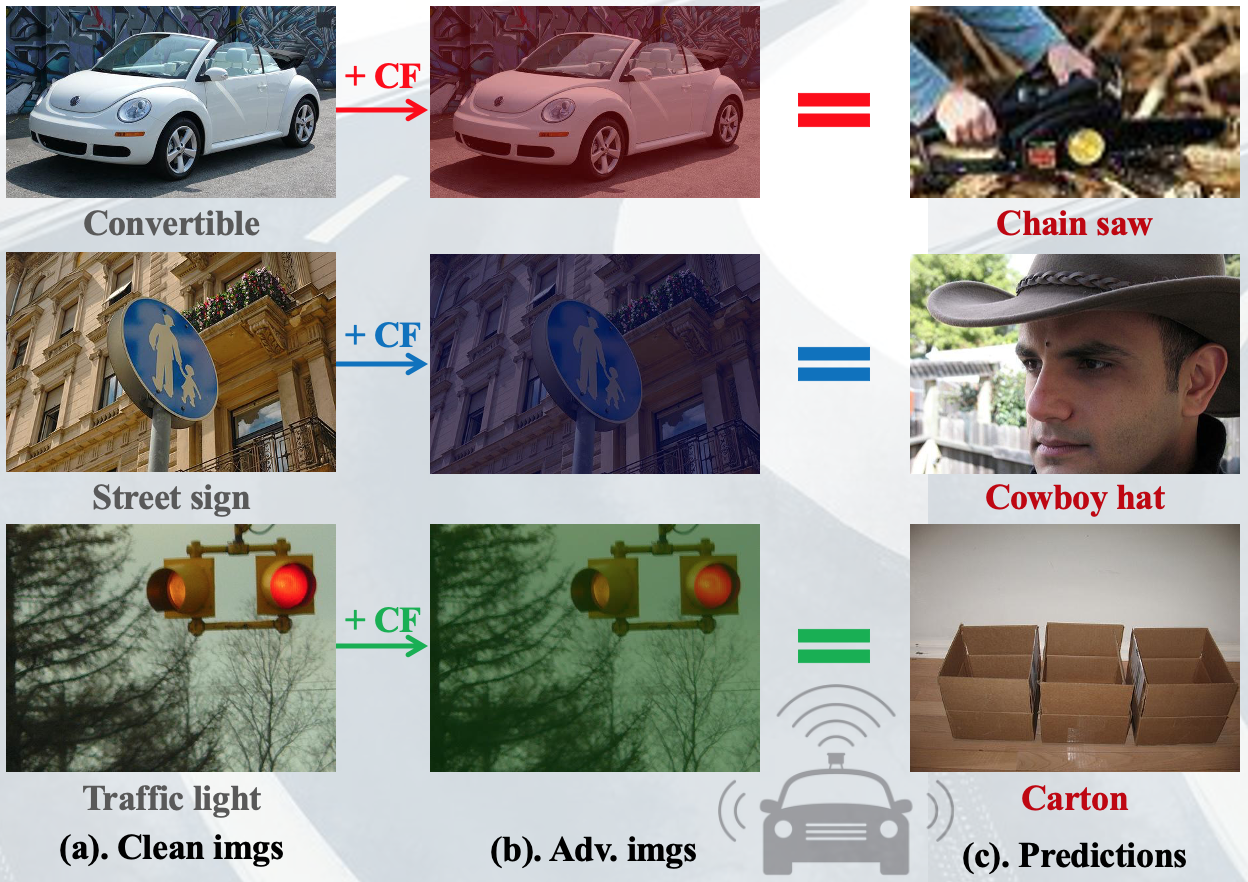} 
\caption{An example. When the camera of an autonomous car is interfered with color film, it fails to recognize convertible, street sign, etc.}.
\label{figure1}
\end{figure}

Many natural factors can be major contributors to physical perturbations. For example, in hazy days with poor visibility, colored haze may be the culprit for self-driving traffic accidents. If a translucent color film is deliberately placed in front of the lens of a self-driving car's camera to create a haze effect on the image to disturb advanced classifiers, it could have car accidents. As shown in Figure \ref{figure1}, the attacker places the optimized color film in front of the camera of the autonomous vehicle, so that the autonomous vehicle could not correctly identify the target objects, carry out malicious attacks and disrupt traffic order.

At present, most physical attacks use stickers as perturbations \cite{ref24,ref26}, which successfully fool advanced DNNs without changing the semantic information of the target object. However, sticker-based attacks are hard to hide. Some scholars use light beam as perturbations \cite{ref35} to execute instantaneous attacks, which successfully fools advanced DNNs as well as achieves better stealthiness. However, light-based attacks are prone to paralysis during the daytime. Some scholars have studied camera-based adversarial attack \cite{ref38}, in which a tiny patch is attached in the camera lens to generate adversarial samples. However, it has high requirements on physical operation and is difficult to adapt to complex attack scenarios.

\begin{figure}
\centering
\setlength{\belowcaptionskip}{-0.3cm}
\includegraphics[width=1\columnwidth]{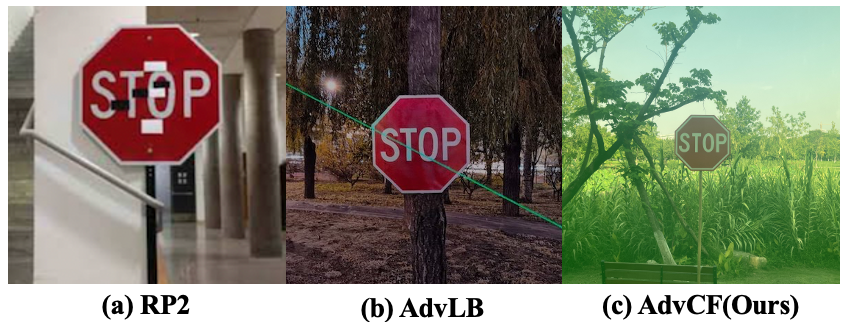} 
\caption{Visual comparison.}.
\label{figure2}
\end{figure}

In this paper, we propose a simple and efficient camera-based physical adversarial attack called Adversarial Color Film (AdvCF). Unlike most existing physical attacks, ours conduct effective physical attacks without modifying the target objects. In terms of stealthiness, a visual comparison of our proposed method with other works is shown in Figure \ref{figure2}. The adversarial samples generated by AdvCF like photos taken on hazy days or when the camera is out of focus. Though the adversarial sample generated by AdvCF may appear less stealthiness than RP2 and AdvLB, AdvCF may exhibit flexible color changes to adapt to various environments. For example, unlike the existing physical attacks focus on a single environmental condition: Sticker-based is applicable to daytime (see RP2 in Figure \ref{figure2})  and light-based is applicable to nighttime (see AdvLB in Figure \ref{figure2}), ours focus on both daytime and nighttime environments (see Figure \ref{figure8}).

Our method is simple to implement physical attacks. By formalizing the physical parameters of color film, using genetic algorithm \cite{ref46} to find the most aggressive physical parameters. Then, based on these physical parameters, printing color films and place them in front of a camera to generate physical samples. We conduct comprehensive experiments to verify the effectiveness of AdvCF. We achieve an 95.9\% attack success rate on a subset of ImageNet in the digital environment. In the physical environment, the attack success rate of indoor test and outdoor test is 80.8\% and 86.7\%, respectively. Furthermore, we use the adversarial samples generated by AdvCF to execute the transfer attack, and verify the AdvCF’s performance in black-box setting. Our main contributions are as follows:

\begin{itemize}
\item We propose a camera-based physical-world attack, AdvCF, which performs efficient physical attacks by manipulating the physical parameters of color film without modifying the target object. At the same time, deploying such attack is very simple: by using an adversarial color film, it could be a common safety threat due to its ease and convenience (See Section \ref{sec1}).
\item We introduce and analyze the existing methods (See Section \ref{sec2}), then, design strict experimental method and conduct comprehensive experiments to verify the effectiveness of AdvCF (See Section \ref{sec3}, Section \ref{sec4}).
\item We conduct a comprehensive analysis of AdvCF, including transfer attacks of AdvCF, defense strategy of AdvCF, etc. These studies will help scholars explore camera-based physical attacks, and enlighten the thinking of defense (See Section \ref{sec5}). At the same time, we look into some promising mentality for camera-based physical attacks (See Section \ref{sec6}).
\end{itemize}

\section{Related work}
\label{sec2}
\subsection{Digital attacks}
Adversarial attack was first proposed by Szegedy et al. \cite{ref1}, and then, adversarial attack was successively proposed \cite{ref16,ref17,ref20,ref21}. Many scholars are committed to the study of adversarial attack in the digital environment.

Most digital attacks generate adversarial perturbations that are bounded by a small norm-ball to ensure imperceptible to human observer. Among them, ${L}_{2}$ and ${L}_{\infty}$ are the most commonly used norms \cite{ref2,ref3,ref4,ref5,ref6}, which guarantee the effectiveness of attacks as well as achieve stealthiness. In addition, some scholars modify other attributes of digital images to generate adversarial samples, for example, color \cite{ref7,ref8,ref9}, texture and camouflage \cite{ref10,ref11,ref12,ref13}, etc. These methods generate perturbations that are slightly perceptible to the human observers. At the same time, some scholars modify the physical parameters of digital images \cite{ref14,ref15} and only retain the key components of images to generate adversarial samples. In general, the assumption of digital attacks is that an attacker can modify the input images, but this is not practical in a physical scenario.

\begin{figure*}
\centering
\setlength{\belowcaptionskip}{-0.5cm}
\includegraphics[width=0.8\linewidth]{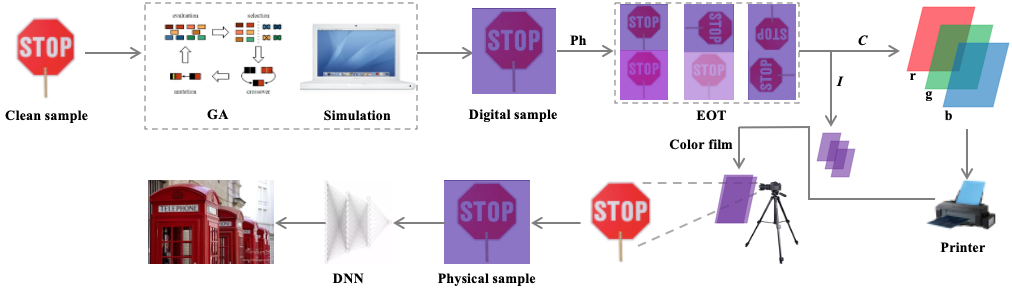}
\caption{Generating an adversarial sample.}.
\label{figure3}
\end{figure*}

\begin{figure}
\centering
\setlength{\belowcaptionskip}{-0.5cm}
\includegraphics[width=0.8\columnwidth]{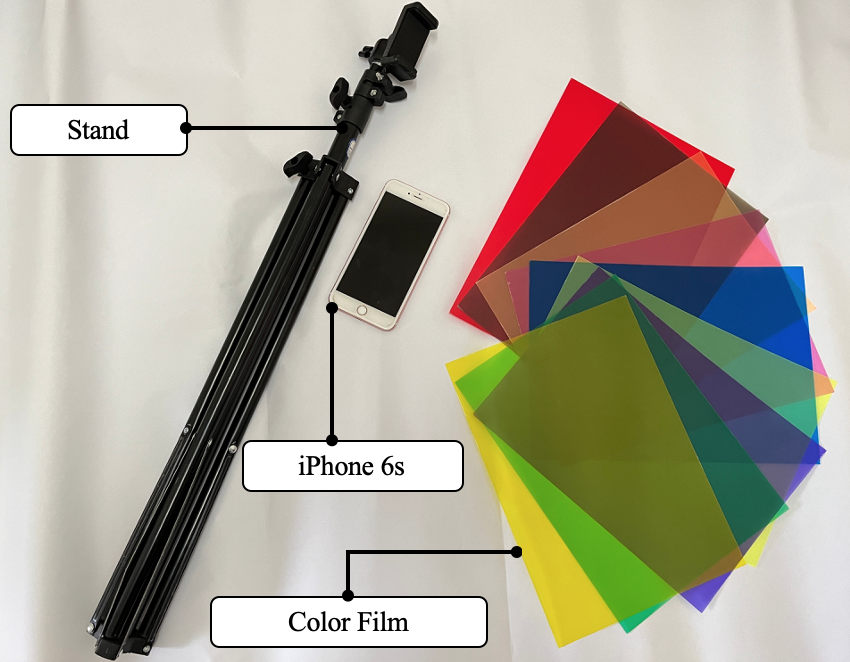}
\caption{Experimental devices.}.
\label{figure5}
\end{figure}

\subsection{Physical attacks}
Physical attack was first proposed by Alexey Kurakin et al. \cite{ref22}. After this work, many physical attacks were proposed successively \cite{ref24,ref28,ref29,ref30,ref31}.

\textbf{Traditional street sign attacks.} Ivan Evtimov et al. \cite{ref24} proposed a classic physical attack called RP2, which uses stickers as perturbations to perform attacks at different distances and angles against advanced DNNs. However, RP2 is susceptible to environmental interference at large distances and angles. Eykholt et al. \cite{ref26} implemented a disappear attack by improving RP2, generating robust and transferable adversarial samples to fool advanced DNNs. However, the perturbations cover a large area, which is too conspicuous. Chen et al. \cite{ref23} proposed ShapeShifter, by using "Expectation over Transformation" to generate adversarial samples, the experimental results showed that the generated stop sign always fooled the advanced DNNs at different distances and angles. Huang et al. \cite{ref27} improved ShapeShifter by adding Gaussian white noise to ShapeShifter's optimization function, achieving a more comprehensive attack. However, ShapeShifter and the improved ShapeShifter have a defect, perturbations cover almost the whole road sign, failed to achieve stealthiness. Duan et al. \cite{ref25} proposed AdvCam, which uses style transfer techniques to generate adversarial samples and disguise the perturbations as a style considered reasonable by human observers. AdvCam has better stealthiness than above methods, but it needs to manually select the attack area and target. All in all, the above methods require manual modification of the target objects. In addition, these works failed to achieve the stealthiness. %Under the influence of complex environment, it is valuable to realize simple, efficient and covert physical attacks.

\textbf{Light-based attacks.} Light-based attacks show some advantages over traditional street sign attacks. Nguyen et al. \cite{ref33} proposed to use projector to attack face recognition system, using light projection as perturbations to generate adversarial samples, verified its antagonism to face recognition system under white-box and black-box settings. However, its deployment mode is complex. Some utilized visible and invisible light to attack face recognition systems \cite{ref32,ref34}, project optimized light onto the target to perform covert attacks. These attacks achieve better stealthiness, but modify the target objects. Duan et al. \cite{ref35} proposed AdvLB, which uses laser beam as perturbations and manipulates its physical parameters to execute attacks. Gnanasambandam et al. \cite{ref36} proposed OPAD, which projects digital perturbations onto the target objects to perform efficient attacks. AdvLB and OPAD, however, can only perform attacks in weak-light conditions. Zhong et al. \cite{ref37} studied shadow-based physical attack, which cast carefully crafted shadows on the target to generate adversarial samples, realizing a natural black-box attack. However, this method is difficult to work in complex physical scenes. To sum up, light-based attacks achieve better stealthiness, but each attack has limitations, including a similar disadvantage: modifying the target objects.

\textbf{Camera-based attacks.} Li et al. \cite{ref38} studied camera-based attacks by placing well-designed stickers on the camera lens to generate adversarial samples, performing targeted attacks against advanced DNNs, it avoids modifying the target by physically manipulating the camera itself, at the same time, adversarial perturbations are inconspicuous. However, it's difficult to adjust error due to its complex operation. Our proposed method puts a carefully designed color film in front of the camera lens, which is simple and efficient to implement in physical scenarios.
%-------------------------------------------------------------------------
\section{Approach}
\label{sec3}
\subsection{Adversarial sample}
Given an input picture $X$, ground truth label $Y$, the DNN classifier  $f$, $f(X)$ represents the classifier's prediction label for picture $X$, The classifier $f$ associates with a confidence score ${f}_{Y}(X)$ to class $Y$. The adversarial sample ${X}_{adv}$ satisfies two properties:  (1) $f({X}_{adv}) \neq f(X) = Y$; (2) $\parallel {X}_{adv} - X \parallel < \epsilon$. Among them, the first property requires ${X}_{adv}$ fools DNN classifier $f$. The second property requires that the perturbations of ${X}_{adv}$ are small enough to be imperceptible to human observers.

In this paper, we use genetic algorithm \cite{ref46} to optimize the physical parameters of color film, and print physical color film according to the physical parameters. Then, in the real scenarios, we put color film in front of the camera lens to take an image and generate an adversarial sample. Figure \ref{figure3} shows our approach.

\subsection{Color film definition}
In this paper, we define a color film using two physical parameters: color $\mathcal{C}(r, g, b)$, intensity $\mathcal{I}$. Each parameter is described as follows:

\textbf{Color $\mathcal{C}(r, g, b)$:}  $\mathcal{C}(r, g, b)$ represents the color of the color film, where $r$, $g$, and $b$ represent the red channel, green channel, and blue channel of the color film respectively.

\textbf{Intensity $\mathcal{I}$:} $\mathcal{I}$ indicates the transparency of color film, the greater $\mathcal{I}$ indicates the lower transparency, the smaller the more transparent. In physical environment, color film with high strength could be generated by superimposing color films.

The parameters $\mathcal{C}(r, g, b)$ and $\mathcal{I}$ form a color film’s physical parameter $Ph(\mathcal{C}, \mathcal{I})$. We define a simple function $S(X, Ph(\mathcal{C}, \mathcal{I}))$ that simply synthesizes the input image with color film to generate an adversarial sample, We define the restriction vectors ${\vartheta}_{min}$ and ${\vartheta}_{max}$ to limit the range of the physical parameters $\mathcal{C}$, and $\mathcal{I}$. The restriction vectors ${\vartheta}_{min}$ and ${\vartheta}_{max}$ are adjustable. Therefore, the adversarial sample can be expressed as:

\begin{equation}
    \label{Formula 1}
    {X}_{adv} = S(X, Ph(\mathcal{C}, \mathcal{I}))
\end{equation}

$$s.t. \quad Ph(\mathcal{C}, \mathcal{I}) \in ({\vartheta}_{min},{\vartheta}_{max})$$

% Function \ref{Formula 1} represents generating an adversarial sample. In next section, we describe how to use genetic algorithm to optimize the physical parameters of color film.

\subsection{Genetic algorithm (GA)}
GA \cite{ref46} is a natural heuristic algorithm designed by John Holland according to the laws of biological evolution in nature. It’s a computational model that simulates the biological evolution process of natural selection and genetic mechanism of Darwin's biological evolution, searches the optimal solution by simulating the natural evolution process. %The flow chart of GA is shown in Figure \ref{figure4}.

% \begin{figure}
% \centering
% \includegraphics[width=0.7\columnwidth]{figures/fig4.png}
% \caption{GA flow chart.}.
% \label{figure4}
% \end{figure}

In this work, we use no model’s gradient information, require only confidence score and prediction label from the model feedback. The feasibility of using GA to optimize AdvCF include:

(1) GA searches the string set of solutions of the problem, covering a wide area, which is conducive to global optimization. In our method, physical parameters $\mathcal{C}$ and $\mathcal{I}$ include a total of $256\times256\times256\times4$ combinations of problem solutions, GA is conducive to the global optimization of AdvCF.

(2) GA basically need no knowledge of search space or other auxiliary information, uses the fitness value to evaluate individuals, and carries out genetic operation on this basis. The fitness function is not constrained by continuous differentiability, and its definition domain can be set arbitrarily. AdvCF does not need model’s gradient information, takes the model’s confidence score ${f}_{Y}(X)$ as the individual fitness, $f({X}_{adv}) \neq Y$ as the termination condition.

(3) Flexible selection strategy. GA uses evolutionary information to organize search. Individuals with high fitness have higher survival probability, and obtain a more adaptable gene structure. AdvCF uses the flexibility of genetic algorithm to select specific elimination strategy to further expand the search scope and achieve global optimization.

We choose binary encoding to encode the physical parameters. For the physical parameters $\mathcal{C}(r, g, b)$, $r$, $g$ and $b$ range from 0 to 255, so $r$, $g$, $b$ correspond to 8 genes respectively. We set intensity $\mathcal{I}$ to four different intensity values (e.g., 0.3 to 0.6) corresponding to 2 genes. Thus, parameters $\mathcal{C}(r, g, b)$ and $\mathcal{I}$ contain a total of 26 genes. After randomly encoding the initial population, utilizing binary conversion to convert genotype into phenotype. For example: 
$Genotype (10010111,00011001,01011101,10)
\rightarrow Phenotype (\mathcal{C}(151,25,93), \mathcal{I}=0.5)$.
Then, input phenotype parameters into the model and generate adversarial samples according to Function \ref{Formula 1}. For the selection strategy, we select from small to large according to the confidence score, and eliminate the individuals with high confidence scores (for example, 1/10). Note that in this work, the smaller the confidence score, the stronger the individual fitness. Crossover and mutation strategies follow the conventional approach. During population iterations, an individual satisfies $f({X}_{adv}) \neq Y$, saving the adversarial sample and physical parameters. In the physical environment, we print the color film according to the physical parameters. place the color film in front of the camera lens to take pictures according to the method shown in Figure \ref{figure3}, and generate physical adversarial samples.

\textbf{Expectation Over Transformation.} EOT \cite{ref31} is an effective tool for handling the conversion from digital to physical domains. We define a transformation $\mathcal{T}$ to represent the domain transition, $\mathcal{T}$ is a random combination of digital image processing, including brightness adaptation, position offset, color variation, and so on. Through EOT, the physical sample can be represented as:

\begin{equation}
    \label{Formula 2}
    {X}_{phy} = \mathcal{T}({X}_{adv}, Ph(\mathcal{C}, \mathcal{I})) 
\end{equation}

$$s.t. \quad Ph(\mathcal{C}, \mathcal{I}) \in ({\vartheta}_{min},{\vartheta}_{max})$$

\begin{figure*}
\centering
\setlength{\belowcaptionskip}{-0.5cm}
\includegraphics[width=1\linewidth]{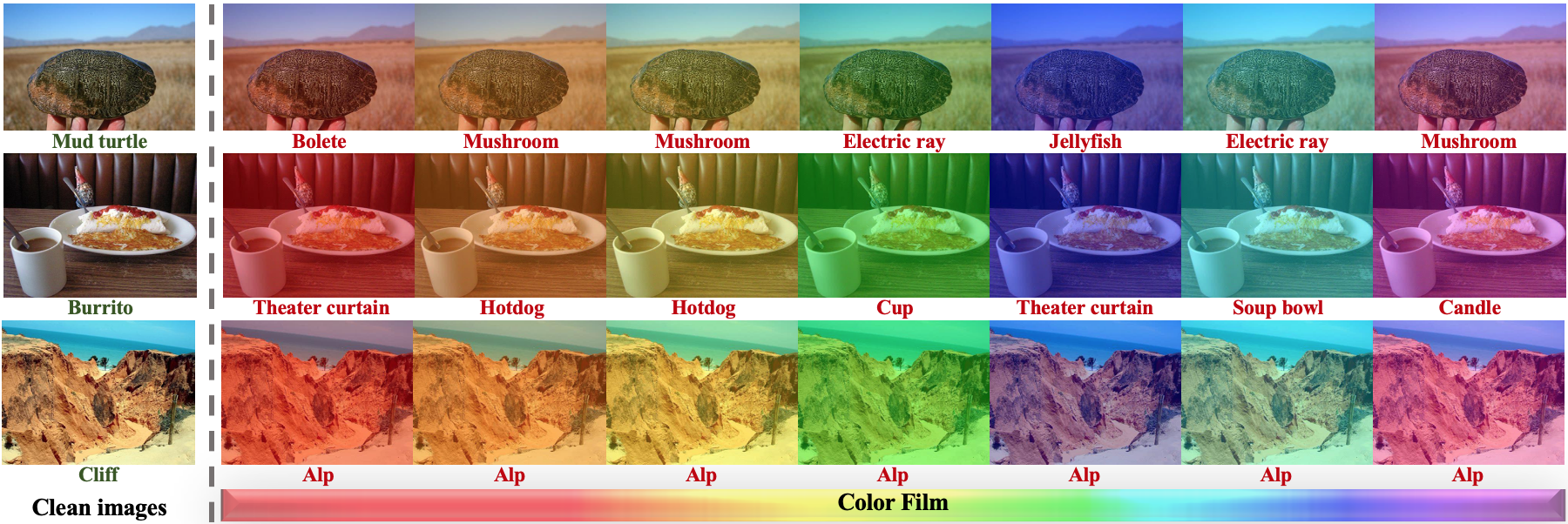}
\caption{Adversarial samples generated by AdvCF.}.
\label{figure6}
\end{figure*}

\begin{figure}
\centering
\setlength{\belowcaptionskip}{-0.5cm}
\includegraphics[width=1\columnwidth]{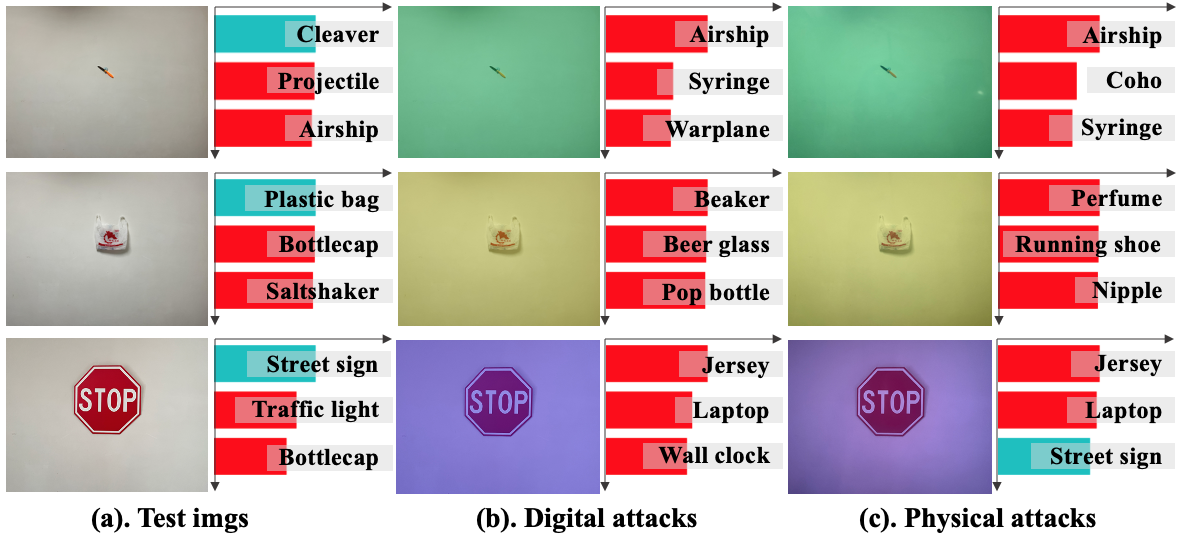}
\caption{Indoor test.}.
\label{figure7}
\end{figure}

\subsection{Color film adversarial attack}
AdvCF focuses on searching $Ph(\mathcal{C}, \mathcal{I})$, the physical parameters of color film, which generates an adversarial sample ${X}_{adv}$ that fools the classifier $f$. In this experiment, we consider a practical situation: the attacker cannot obtain the knowledge of the model, but only the confidence score ${f}_{Y}(X)$ with given input image $X$ on ground truth label $Y$. In our proposed method, we use confidence score as the adversarial loss. Thus, the objective is formalized as minimizing the confidence score on the ground truth label $Y$, as is shown in follows:

\begin{equation}
    \label{Formula 3}
    \mathop{\arg\min}_{Ph}{\mathbb{E}}_{t \sim \mathcal{T}}[{f}_{Y}(t({X}_{adv}, Ph(\mathcal{C},  \mathcal{I})))]
\end{equation}

$$ s.t. \quad f({X}_{adv}) \neq Y$$

\begin{figure*}
\centering
\setlength{\belowcaptionskip}{-0.5cm}
\includegraphics[width=1\linewidth]{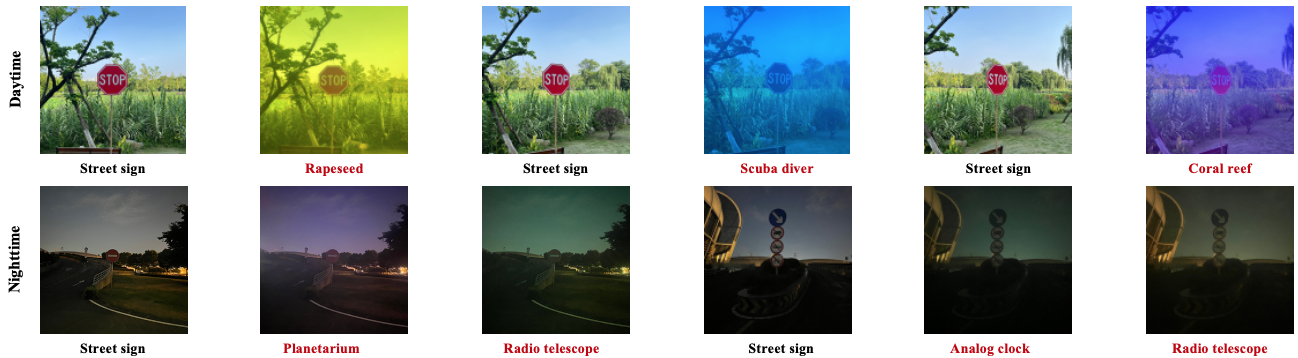}
\caption{Outdoor test.}.
\label{figure8}
\end{figure*}

Algorithm \ref{algorithm1} shows the pseudocode of AdvCF. The proposed method takes clean image $X$, target classifier $f$, correct label $Y$, population $Seed$, iteration number $Step$, crossover rate $Pc$, mutation rate $Pm$ as input decided by the attacker. Then, binary encoding the initial population of color film. Within the max iterations, caculating the fitness value of each individual in the population. If an individual attacks the target model successfully, output its physical parameters and terminate the algorithm. In each iteration, the selection stradegy follows the fitness value ${f}_{Y}({X}_{adv})$, taking $Pc$ and $Pm$ for crossover and mutation, respectively. Here, our selection strategy is to weed out the top tenth with the largest confidence score (note that the smaller the confidence score, the more antagonistic) and then fill in the randomly encoded genes separately. The benefit of this selection strategy is that it saves a lot of time cost by directly eliminating the most inferior individuals, and further expand the search scope and global optimization. In addition, we set crossover rate $Pc$ and variation rate $Pm$ to 0.7 and 0.1, respectively. Experiments in Figure  \ref{figure10} show that our strategies for selection, crossover and mutation perform efficient optimization on the target problem. The algorithm finally outputs the physical parameters of the color film ${Ph}^{\star}$, which are used to perform subsequent physical attacks.

\begin{algorithm}
	\renewcommand{\algorithmicrequire}{\textbf{Input:}}
	\renewcommand{\algorithmicensure}{\textbf{Output:}}
	\caption{Pseudocode of AdvCF}
	\label{algorithm1}
	\begin{algorithmic}[1]
	
		\REQUIRE Input $X$, Classifier $f$, Ground truth label $Y$, Population size $Seed$, Iterations $Step$, Crossover rate $Pc$, Mutation rate $Pm$;
		\ENSURE A vector of parameters ${Ph}^{\star}$;

		\STATE \textbf{Initialization} $Seed$, $Step$, $Pc$, $Pm$;
		
		\FOR{$seeds$ $\leftarrow$ 0 to $Seed$}
		    \STATE Encoding individual genotype ${G}_{seeds}$;
		\ENDFOR
		
		\FOR{$steps$ $\leftarrow$ 0 to $Step$}
		\FOR{$seeds$ $\leftarrow$ 0 to $Seed$}
		    \STATE ${Ph}_{seeds}(\mathcal{C}, \mathcal{I}) \leftarrow{{G}_{seeds}}$;
		    \STATE ${X}_{adv}(seeds) = S(X, {Ph}_{seeds}(\mathcal{C}, \mathcal{I}))$;
		    \STATE ${f}_{Y}({X}_{adv}) \leftarrow f({X}_{adv})$;
		    
		    \IF{$f({X}_{adv}) \neq Y$}
		        \STATE ${Ph}^{\star} = {Ph}_{seeds}(\mathcal{C}, \mathcal{I})$;
		        \STATE \textbf{output} ${Ph}^{\star}$;\\
		        \STATE break;
		    \ENDIF
		\ENDFOR
		    \STATE Update: ${G}_{seed}\xleftarrow{Selection}{f}_{Y}({X}_{adv})$;
            \STATE Update: ${G}_{seed}\xleftarrow{Crossover} Pc$;
            \STATE Update: ${G}_{seed}\xleftarrow{Mutation} Pm$;
		\ENDFOR
		
	\end{algorithmic}  
\end{algorithm}

%------------------------------------------------------------------------

\section{Evaluation}
\label{sec4}
\subsection{Experimental setting}
We test the effectiveness of AdvCF in both digital and physical environments. We perform all experiments using ResNet50 \cite{ref40} as the target model. As with the method in AdvLB \cite{ref35}, we randomly selected 1000 images in ImageNet \cite{ref47} that could be correctly classified by ResNet50 as a dataset for the digital test. In physical test, the experimental devices are shown in Figure \ref{figure5}. We show part of the color films and use iPhone6s as the camera device. It has been verified that different camera devices will not affect the effectiveness of AdvCF. For all tests, we use attack success rate (ASR) as the metric to report the effectiveness of AdvCF.

\subsection{Evaluation of AdvCF}
\textbf{Digital test.} We test the effectiveness of AdvCF in digital environments on 1000 images that could be correctly classified by ResNet50, achieving an attack success rate of 95.90\% (Untargeted ASR of 95.10\% in AdvLB \cite{ref35}, targeted ARS of 49.60\% in \cite{ref38}). Figure \ref{figure6} shows some interesting results. For example, when adding a blue ($\mathcal{C}(0, 0, 255)$) film to the clean sample, the mud turtle is misclassified as jellyfish, a similar phenomenon can be seen in \cite{ref35}. Studies have shown that most of the original jellyfish-label samples in the ImageNet training set are with blue backgrounds, so adding blue film will prone to lead the classifier to misclassify the adversarial samples into jellyfish, which is described in \cite{ref49}. On the other hand, the cliff is misclassified as alp when covering various color films. All in all, AdvCF shows an effective adversarial effect in the digital environment, that is, it leads advanced DNNs to misclassification without changing the semantic information of the target objects.

\textbf{Physical test.} To demonstrate the rigor of AdvCF, we conduct a strict experimental design in the physical test. In the physical world, the effectiveness of physical attack is affected by environmental noise, so we design indoor test and outdoor test respectively. In which, the indoor test avoids the influence of outdoor noise, and the outdoor test reflects the performance of AdvCF in real scenarios.

For the indoor test, we use ‘Cleaver’, ‘Plastic bag’, ‘Street sign’, etc. as target object. and form 26 adversarial samples, achieving an attack success rate of 80.77\% (ASR of 100\% in AdvLB \cite{ref35}). Figure \ref{figure7} shows the experimental results of the indoor test. It can be seen that the computer-simulated color films keep better consistency with the physical printed color films. In addition, there exists printing loss of the physical printed color film due to the limitations of current printing equipment, which could be evaded by EOT \cite{ref31}.

\begin{table}
    \setlength{\belowdisplayskip}{-0.5cm}
    \centering
    \caption{\label{Table 1}Attack success rates from different angles.}
    \begin{tabular}{cccc}
    \hline
    \quad & ${0}^{\circ}$ & ${30}^{\circ}$ & ${45}^{\circ}$ \\
    \hline
    ASR (\%) & 85.19 & 77.78 & 85.19\\
    \hline
    \end{tabular}
\end{table}

In the outdoor test, we select ‘Black swan’, ‘Ashcan’ and ‘Street sign’ as attack objects, and form 105 adversarial samples, achieving an attack success rate of 86.67\% (ASR of 77.43\% in AdvLB \cite{ref35}, ARS of 73.26\% in \cite{ref38}). Figure \ref{figure8} shows adversarial samples in the outdoor environment. Experimental result shows that by adding the optimized color film interference to the clean samples, it leads the advanced DNNs to misclassification. On the other hand, to get close to real scenarios, we conduct outdoor tests on ‘Stop sign’ from different angles, and the experimental results are shown in Table \ref{Table 1}. It shows that AdvCF performs effective physical attacks on target objects at various angles.

\begin{table*}[htbp]
    \centering
    \caption{\label{Table 2}Ablation study of $\mathcal{I}$.}
    \begin{tabular}{cccccccc}
    \hline
    $\mathcal{I}$ & 0.1 & 0.2 & 0.3 & 0.4 & 0.5 & 0.6 & 0.7 \\
    \hline
    ASR (\%) & 15.80 & 27.60 & 38.90 & 54.60 & 69.20 & 82.10 & 92.40\\
    \hline
    \end{tabular}
\end{table*}

In general, AdvCF avoids directly modifying the target objects. Besides, the proposed method is capable of performing physical attacks during daytime and nighttime, compared to light-based attacks and sticker-based attacks. In addition, while our approach does not perform as well as AdvLB in indoor test, it shows better adversarial effect in digital test and outdoor test and is more adaptable to various scenarios than AdvLB. In summary, the comprehensive experimental results show that AdvCF is effective in both digital and physical environments.

\subsection{Ablation study}
Here, we perform a series of experiments to study the adversarial effects of different physical parameters on AdvCF: Intensity $\mathcal{I}$; Color $\mathcal{C}(r, g, b)$.

\textbf{Intensity $\mathcal{I}$:} Here, we conduct experiments on 1000 images that could be correctly classified by ResNet50. The greater $\mathcal{I}$, the stronger adversarial effect, and the worse concealment. We study the adversarial effect of color films with intensity of 0.1 to 0.7. Table \ref{Table 2} shows the attack success rates for each intensity of color film. It shows that AdvCF is aggressive even at a weak intensity.

\begin{figure}[H]
\centering
\includegraphics[width=1\columnwidth]{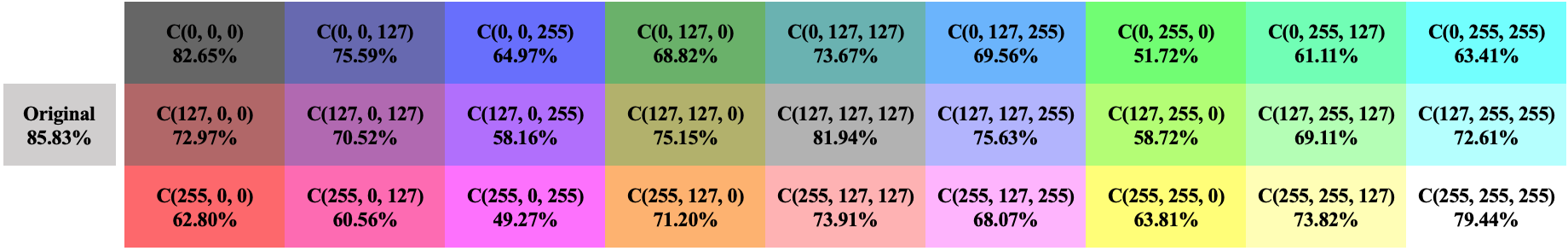}
\caption{Ablation study of $\mathcal{C}(r, g, b)$.}.
\label{figure9}
\end{figure}

\textbf{Color $\mathcal{C}(r, g, b)$:} Here, in order to strictly study the adversarial effect of color films with different colors, we construct a larger dataset. First of all, we randomly selected 50 clean samples from each of the 1000 categories in ImageNet \cite{ref47} and got 50,000 clean samples. Secondly, add 27 color films with $\mathcal{I}=0.4$ to each clean sample to obtain the final data set containing 1.35 million adversarial samples, which is called ImageNet-ColorFilm (ImageNet-CF). Figure \ref{figure9} shows the classification accuracy of resnet50 \cite{ref40} on clean samples and adversarial samples. It can be seen that each color film shows the adversarial effect. Among them, adversarial samples related to $\mathcal{C}(255, 0, 255)$ are the most adversarial, and the classification accuracy of resnet50 is 49.27\% ($\downarrow$36.56\%).

%------------------------------------------------------------------------
\section{Discussion}
\label{sec5}
\subsection{Effectiveness of GA optimization}
\begin{figure}
\centering
\setlength{\belowcaptionskip}{-0.1cm}
\includegraphics[width=1\columnwidth]{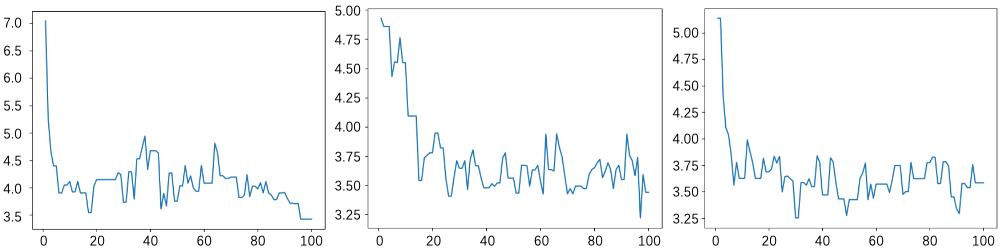}
\caption{GA optimization.}.
\label{figure10}
\end{figure}

We conduct experiments to evaluate GA \cite{ref46}, in which the number of iterations is 100 and the population size is 100. We monitor the optimization process of GA and verify its effectiveness. As shown in Figure \ref{figure10}, the horizontal axis represents 100 iterations, and the vertical axis represents the confidence score of the most adversarial individual in the population after each iteration. The experimental results show that GA leads the objective function to converge in a few iterations, which verifies the effectiveness of GA on AdvCF.

\subsection{Model attention}
\begin{figure}[H]
\centering
\includegraphics[width=1\columnwidth]{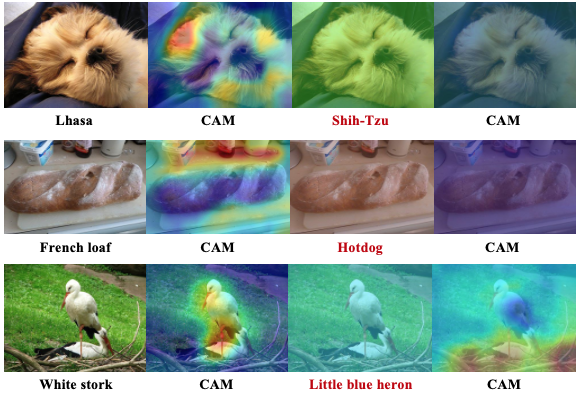}
\caption{CAM for images.}.
\label{figure11}
\end{figure}

We use CAM \cite{ref48} to show the model’s attention. As shown in Figure \ref{figure11}, by adding the optimized color film to clean samples, model’s attention disappears from the target objects or shifts to another one.

\begin{figure*}
\centering
\includegraphics[width=1\linewidth]{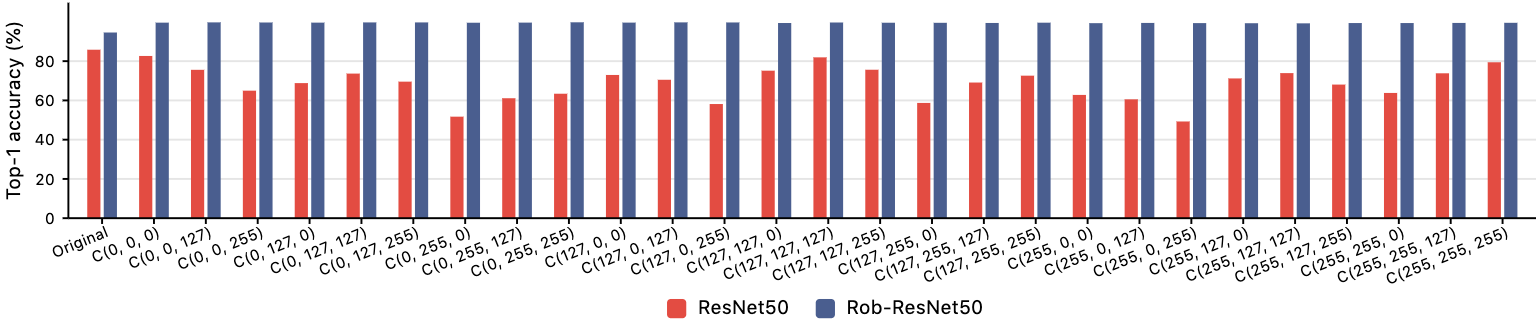}
\caption{ResNet50 vs. Rob-ResNet50.}.
\label{figure12}
\end{figure*}

\subsection{Transferability of AdvCF}
Here, we demonstrate the attack transferability of AdvCF against advanced DNNs \cite{ref45, ref41, ref40, ref42, ref43, ref39, ref44} in both digital and physical environments. We take the adversarial samples generated by AdvCF that successfully attacked resnet50 as the dataset. The experimental results are shown in Table \ref{Table 3}. It can be seen that AdvCF shows effective attack transferability in the digital environment, and the attack success rate against AlexNet is 96.44\%. In the physical environments, AdvCF demonstrates excellent attack transferability, whose black-box attack has paralyzed almost all of the advanced DNNs. Our experimental results imply that AdvCF allows attackers to exploit the transferability of AdvCF to carry out efficient physical attacks against advanced DNNs without any knowledge of the model.

The experimental results in Table \ref{Table 1} show that AdvCF conducts effective physical attacks in a white-box setting. The data in Table \ref{Table 3} show that AdvCF has excellent physical adversarial performance in a black-box setting. AdvCF empowers attacker flexible operations, even without any knowledge of the model, to perform effective physical attacks. Therefore, in view of the excellent adversarial effect of AdvCF on the vision-based system in real scenes, we call for the attention of the proposed AdvCF.

\begin{table}[ht]
\centering
\caption{\label{Table 3}Transferability of AdvCF (ASR (\%)).}
\begin{tabular}{ccccccc}

\hline
\multicolumn{2}{c}{\multirow{2}{*}{$f$}} & \multicolumn{2}{c}{\multirow{2}{*}{Digital}} & \multicolumn{3}{c}{Physical}\\
% \cmidrule[1pt](1r){5-7} %短横线
\cmidrule(r){5-7}

\multicolumn{2}{c}{} & \multicolumn{2}{c}{} & ${0}^{\circ}$ & ${30}^{\circ}$ & ${45}^{\circ}$\\
\hline

\multicolumn{2}{c}{Inception v3} & \multicolumn{2}{c}{26.05} & 80.00 & 92.31 & 86.67\\
\hline
\multicolumn{2}{c}{VGG19} & \multicolumn{2}{c}{60.48} & 93.33 & 92.31 & 100\\
\hline
\multicolumn{2}{c}{ResNet101} & \multicolumn{2}{c}{35.96} & 100 & 92.31 & 100\\
\hline
\multicolumn{2}{c}{GoogleNet} & \multicolumn{2}{c}{28.59} & 100 & 100 & 100\\
\hline
\multicolumn{2}{c}{AlexNet} & \multicolumn{2}{c}{96.44} & 100 & 100 & 100\\
\hline
\multicolumn{2}{c}{DenseNet} & \multicolumn{2}{c}{35.96} & 100 & 100 & 100\\
\hline
\multicolumn{2}{c}{MobileNet} & \multicolumn{2}{c}{67.85} & 100 & 100 & 100\\
\hline

\end{tabular}
\end{table}

\subsection{Defense of AdvCF}
In addition to demonstrating the potential threats of AdvCF, we attempt to defense against AdvCF with adversarial training, and choose the proposed ImageNet-CF as the dataset for adversarial training. We use torchvision to train the ResNet50 robust model (Rob-ResNet50). The model was optimized on 3 2080Ti GPUs by ADAM with initial learning rate 0.01. The experimental results are shown in Figure \ref{figure12}. It can be seen that Rob-ResNet50 achieves a classification accuracy more than 95\% for adversarial samples, which will help scholars to expand adversarial defense strategies against AdvCF in both digital and physical environments.

\subsection{Disadvantages of AdvCF}
We demonstrate the effectiveness of AdvCF in both digital and physical environments.
Here, we summarize some of AdvCF's shortcomings: (1) Although the printing loss is imperceptible to human observers, it actually exists. Due to the limitations of current printing devices, our current precautions can only be carried out in a digital environment. (2) Our experimental studies are not yet able to contribute to the interpretability of DNNs, which will be the direction of our future efforts.

\section{Conclusion}
\label{sec6}
In this paper, we present a simple and efficient camera-based physical-world attack, AdvCF, which performs attacks by manipulating the physical parameters of color film. We summarize and analyze the existing advanced physical attacks. Then, we design strict experimental method and conduct comprehensive experiments to verify the effectiveness of AdvCF. Finally, we analyze the efficiency of our optimization strategy, model’s bias caused by AdvCF, the black-box adversarial effect of AdvCF, the defense guidance against AdvCF, etc. Our experiments reveal the security threats of AdvCF to vision-based applications and systems in the real world. At the same time, deploying AdvCF is rather simple, so that it could be a common safety threat. Our approach provides many ideas for future physical attacks, performing physical operations on the camera itself rather than modifying the target objects. In general, our proposed AdvCF is very beneficial to study the security risks of vision-based applications in real scenarios, it is a valuable complement to the physical-world attacks.

In the future, we will continue to focus on camera-based attacks, such as doodling transparent film, clipping color film to perform attacks, etc. Furthermore, we will deploy camera-based attacks to other computer vision tasks including target detection, domain segmentation. Meanwhile, the defense strategies against camera-based attacks will also become a hot topic in the future.

%%%%%%%%% REFERENCES
{\small
\bibliographystyle{ieee_fullname}
\bibliography{egbib}
}

\end{document}